%% file: paper_v5.tex
\PassOptionsToPackage{hyperfootnotes=false}{hyperref}
\documentclass[11pt,a4paper,logo,copyright,nonumbering]{unimobilebench}
\usepackage[numbers,sort&compress]{natbib}
\usepackage{float}
\usepackage{makecell,multirow,ragged2e}
\usepackage{xspace}
\usepackage{fontawesome5}
\definecolor{citeblue}{rgb}{0.21,0.49,0.74}
\hypersetup{
    pdfborderstyle={/S/S/W 1},
    linkbordercolor={0.4 0.7 1},
    citecolor=citeblue,
    breaklinks=true,
    colorlinks=true,
    linkcolor=red,
    pdftitle={MobilePA-Bench: Benchmarking Mobile Planner Agents on Complex Real-World Tasks},
    pdfauthor={Yi Zhu, Xiongwei Wu, Qiyi Wang, Tingyu Qu, Jiajun Liu, Sihan Cao, Long Chen, Weigao Sun, Feida Zhu, Yiran Zhong, Steven HOI}
}
\newcommand{\ourmodel}{MobilePA-Bench\xspace}
\newcommand{\UniMobileBenchLogo}{\raisebox{-0.18em}{\includegraphics[height=1.1em]{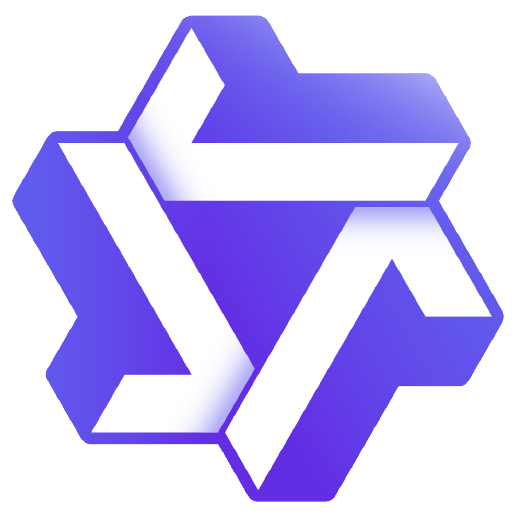}}}

\newsavebox{\arrtablebox}
\newcommand{\fitwide}[1]{%
  \sbox{\arrtablebox}{#1}%
  \ifdim\wd\arrtablebox>\linewidth
    \resizebox{\linewidth}{!}{\usebox{\arrtablebox}}%
  \else
    \usebox{\arrtablebox}%
  \fi}

\title{\centering \ourmodel: Benchmarking Mobile Planner Agents on Complex Real-World Tasks}
\input{authors}

\begin{document}

\begin{abstract}
{Mobile planning requires a central agent to coordinate tools, persistent
user context, reusable procedures, and specialized sub-agents while adapting
to execution feedback. Evaluating these workflows requires distinguishing
task completion from compliance with prescribed resource and procedural
requirements. We introduce \textbf{MobilePA-Bench}, a benchmark comprising
1,705 tasks across eight query task domains and 212 mobile tools in a
stateful simulated environment.
It combines three features: (1)~\textbf{unified capability coverage}, profiling
\emph{Tool Use}, \emph{Memory Usage}, \emph{Skill Usage}, and
\emph{Sub-agent Collaboration} under a common mobile protocol;
(2)~\textbf{evidence-aligned verification}, matching each task to a fixed
tool-call, state-change, or interaction-behavior checker rather than imposing
one completion criterion on all tasks; and (3)~\textbf{capability-gated
evaluation}, requiring task completion alongside prescribed memory checks
and, in skill-only routing, complete route loading before business-tool use.
With MobilePA-Bench, we systematically evaluate open- and closed-weight LLMs,
revealing capability-specific weaknesses in complex mobile task execution.
Fine-grained analyses distinguish failures in resource retrieval and
procedural compliance from downstream execution errors, providing diagnostic
insights beyond aggregate task-success rates. We further train a
Qwen3.8-27B baseline with online reinforcement learning in the sandbox,
whose improved performance provides initial evidence of the environment's
utility for planner development.}

\end{abstract}

\maketitle
\begingroup
\renewcommand{\thefootnote}{}
\footnotetext{\scriptsize\mbox{$^{*}$ Equal contribution.\quad $^{\ddagger}$ Research intern.\quad \textsuperscript{\S} Project lead.}}
\addtocounter{footnote}{-1}
\endgroup

\section{Introduction}
\label{sec:intro}

\begin{figure}[!t]
    \centering
    \includegraphics[width=0.95\linewidth]{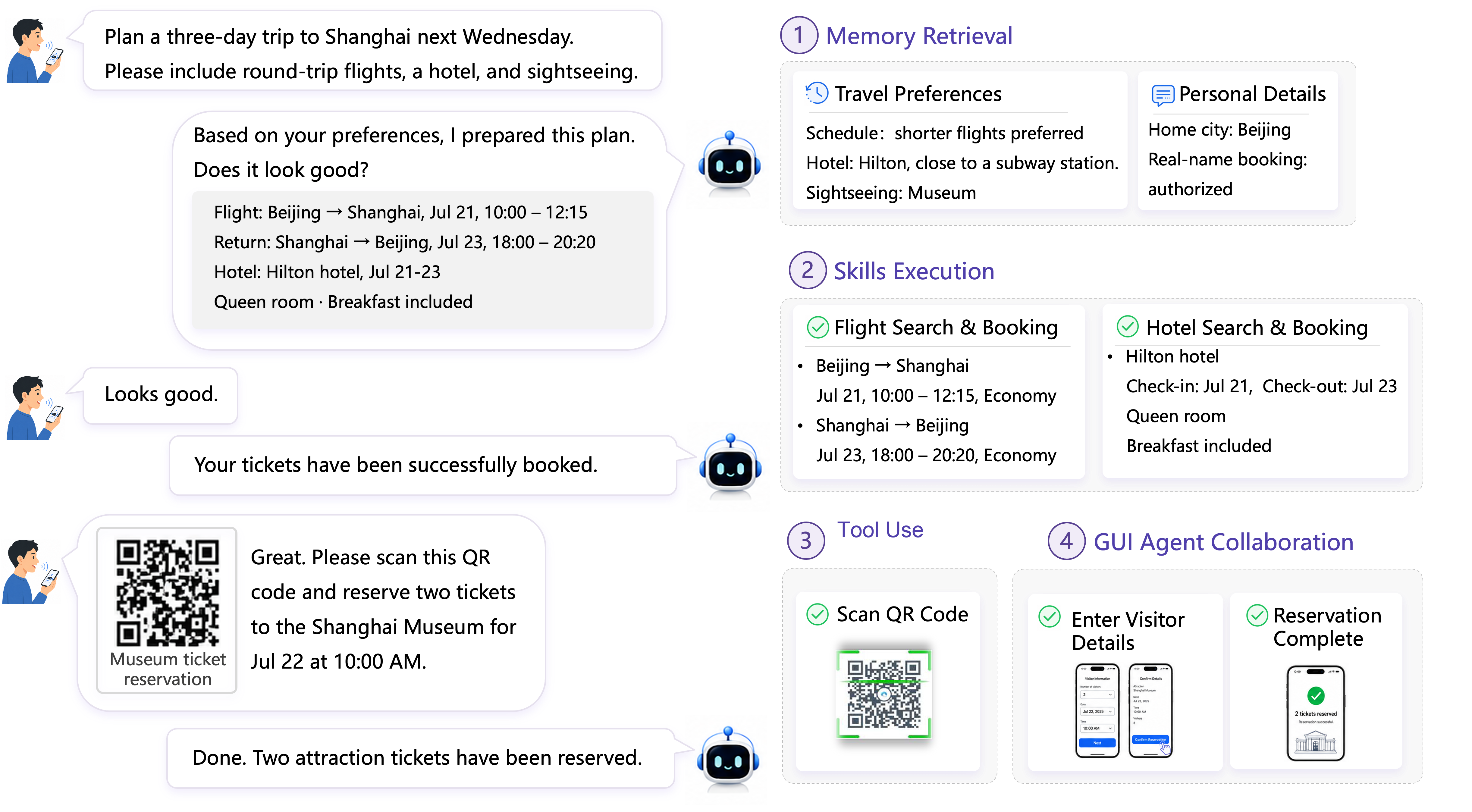}

    \caption{\textbf{An illustrative mobile workflow involving four planner capabilities.} The central planner invokes mobile tools (\textbf{Tool Use}), retrieves user preferences (\textbf{Memory Usage}), loads reusable procedures (\textbf{Skill Usage}), and delegates interface operations to a GUI agent, one of its available sub-agents (\textbf{Sub-agent Collaboration}). }

\label{fig:teaser}
\end{figure}

Mobile agents are evolving from single-step interface operators into personal assistants that handle multi-step tasks across applications. Arranging travel around a user's preferences, for example, may require querying application state, invoking tools, retrieving personal information, following reusable procedures, and delegating operations to specialized sub-agents (Figure~\ref{fig:teaser}). A central \emph{mobile planner} must coordinate these resources, order its actions, and revise its plan from execution feedback. Evaluating mobile planning therefore requires not only measuring task completion, but also identifying weaknesses in the capabilities involved.

Existing benchmarks offer complementary views of agent capabilities. GUI benchmarks assess task completion through application interfaces \citep{rawles2024androidworld,xie2024osworld}; DroidCall evaluates mobile API selection and argument generation \citep{droidcall2024}; and AppWorld, $\tau^3$-bench, and Toolathlon assess interactive, stateful tool execution \citep{trivedi2024appworld,barres2026tau3,li2026toolathlon}. MemoryArena, SkillsBench, and MultiAgentBench examine memory, skills, and multi-agent collaboration, respectively \citep{he2026memoryarena,li2026skillsbench,zhu2025multiagentbench}. Their environments, action interfaces, and success criteria differ. For central mobile planners, we seek a capability profile under a common environment and protocol that connects task outcomes with capability-use evidence. Outcome checks alone cannot establish whether prescribed resource or procedural requirements were met; invocation checks alone cannot establish subsequent task completion.

To this end, we introduce \textbf{MobilePA-Bench}, a stateful, executable benchmark for central mobile planners, comprising 1,705 tasks across eight query task domains and 212 mobile tools. Its \textbf{unified capability coverage} spans four dimensions: \emph{Tool Use} tests tool invocation, argument grounding, and cross-step dependencies; \emph{Memory Usage} examines retrieval, maintenance, and memory-related task execution; \emph{Skill Usage} tests completion under skill-only and mixed tool-skill access; and \emph{Sub-agent Collaboration} evaluates routing, contextual handoff, and follow-up, with GUI agents among the available specialists. These task subsets run through shared interfaces in a simulated environment with persistent application state, operation dependencies, and structured feedback, enabling analysis of overall performance and individual capabilities under a common mobile protocol.

To ground evaluation in inspectable execution evidence, MobilePA-Bench combines \textbf{evidence-aligned verification} with \textbf{capability-gated evaluation}. Each task receives a fixed primary checker during annotation, using tool calls, state changes, or interaction behavior according to its completion semantics. Capability gates check prescribed memory access and action constraints, or timely skill loading in skill-only routing; mixed tool-skill routing also accepts valid direct-tool solutions. Reporting these checks alongside task outcomes separates unmet requirements from downstream execution failures without penalizing valid alternative paths in the mixed setting. The benchmark thus identifies observable failure stages beyond aggregate task-success rates, supporting diagnosis of capability-specific weaknesses in complex workflows.

With MobilePA-Bench, we systematically evaluate open- and closed-weight LLMs on complex mobile tasks. Fine-grained analyses reveal difficulties in argument grounding, dependency planning, and boundary handling. Retrieving required memories or loading skills does not ensure subsequent task completion, while selecting an appropriate sub-agent does not ensure that the full interaction requirements are met. We further train a \texttt{Qwen3.8-27B} baseline with online reinforcement learning in the sandbox and observe improved benchmark performance, providing initial evidence that this evaluation environment can also support planner development.

\section{Related Work}
\label{sec:related-works}

\begin{figure}[!t]
    \centering
    \includegraphics[width=0.95\linewidth]{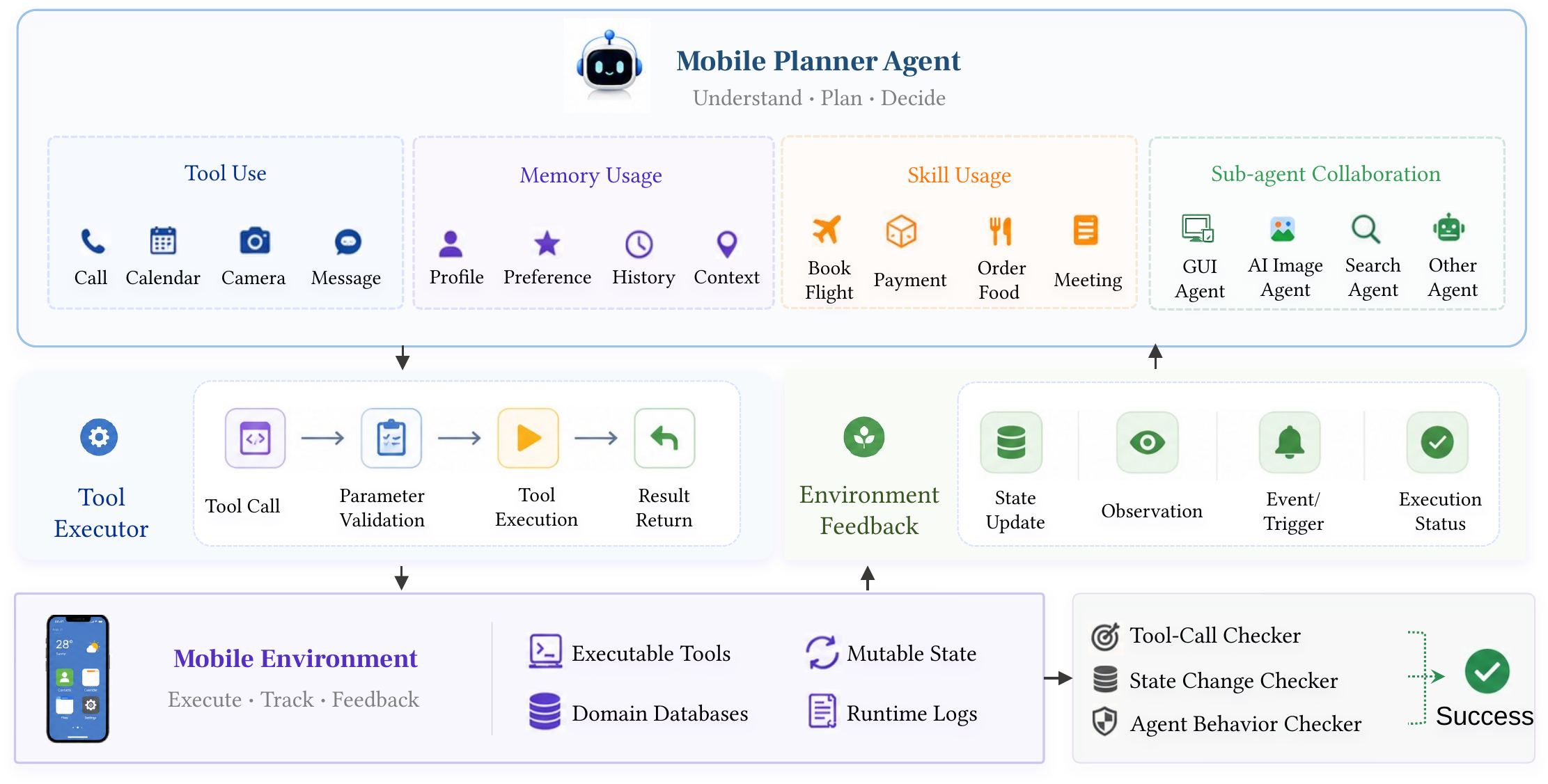}
    \caption{\textbf{Overview of MobilePA-Bench.} A central mobile planner acts through unified interfaces for tools, memory, skills, and specialized sub-agents. The stateful mobile environment executes each action and returns structured feedback, while the assigned verifier evaluates the resulting calls, state changes, or interaction behavior.}

    \label{fig:overview}
\end{figure}

\textbf{Mobile and Tool-Using Agent Benchmarks}
Agent benchmarks cover GUI-based task completion \citep{rawles2024androidworld,xie2024osworld}, mobile API-call prediction \citep{droidcall2024}, and interactive tool execution \citep{trivedi2024appworld,barres2026tau3,li2026toolathlon}. LlamaTouch evaluates on-device execution through essential application states \citep{zhang2024llamatouch}; VitaBench studies multi-turn tool use and changing user intent in service simulations \citep{he2025vitabench}. MobilePA-Bench instead isolates central mobile planning through structured tools and delegation in a stateful sandbox.

\textbf{Memory, Skills, and Agent Delegation}
LongMemEval and BFCL v4 assess memory \citep{wu2025longmemeval,mao2025bfclmemory}, while MemoryArena evaluates memory acquisition and reuse across interdependent agentic sessions \citep{he2026memoryarena}. SkillsBench measures skill augmentation, and MultiAgentBench evaluates collaboration \citep{li2026skillsbench,zhu2025multiagentbench}. MobilePA-Bench combines mobile-task completion with annotated resource-access and procedural requirements. Its gates measure observable compliance, not the causal contribution of resource contents; its Sub-agent score concerns routing and handoff rather than downstream execution.

\textbf{Task Verification and Evaluation}
AppWorld and $\tau$-bench verify database outcomes \citep{trivedi2024appworld,yao2025taubench}, while BFCL uses call- and state-based checks \citep{mao2025bfclmemory}. MCPAgentBench evaluates tool selection with distractors and simulated MCP tools \citep{liu2025mcpagentbench}. MobilePA-Bench combines task-specific call, state, or behavior checkers with capability gates.

\section{MobilePA-Bench}
\label{sec:framework}

MobilePA-Bench separates three evaluation questions: which planner capabilities
are covered, what evidence establishes task completion, and which
capability-specific conditions must also hold. A shared executable environment
supports unified capability coverage; evidence-aligned checkers assess
completion; and capability gates enforce applicable memory and skill-only
routing requirements. Figure~\ref{fig:overview} summarizes the execution framework.

\subsection{Executable Mobile Environment}
\label{sec:benchmark-settings}
\label{sec:sandbox}
\label{sec:agent-harness}

The stateful mobile simulator (Figure~\ref{fig:overview}) evaluates central-planner routing and cross-step coordination, with a GUI agent available as one specialized sub-agent. It preserves application records, operation dependencies, and structured failures; visual perception, UI dynamics, and device-level execution are outside its scope.

\begin{figure}[!t]
    \centering
    \includegraphics[width=0.84\linewidth]{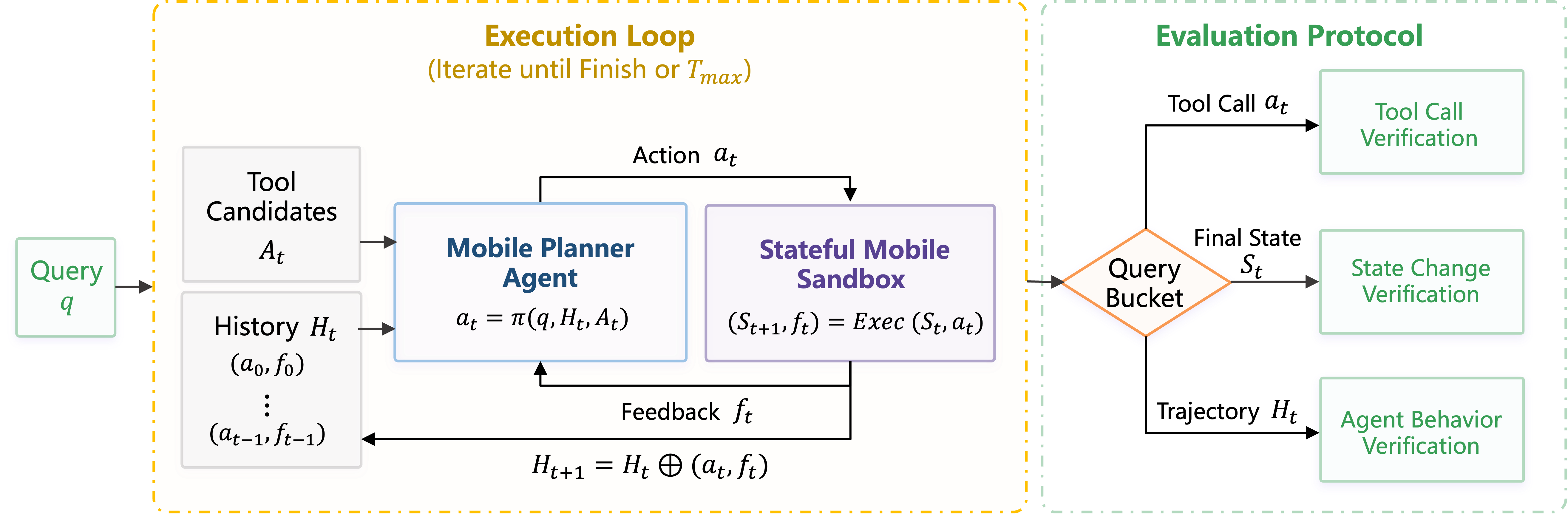}
    \caption{\textbf{Closed-loop execution and verification.} The planner acts, the sandbox updates state and returns feedback, and the assigned checker evaluates the terminal trajectory and state.}
    \label{fig:evaluation-pipeline}
\end{figure}

\textbf{Task Formulation.}
A mobile task is $\tau=(q,\mathcal{S}_0,\mathcal{H}_0,\mathcal{A}_0)$, where $q$ is the user request, $\mathcal{S}_0$ the initial state, $\mathcal{H}_0$ the optional dialogue history, and $\mathcal{A}_0$ the active candidate action set. At step $t$, the planner acts and the sandbox updates the state and public interaction history:
\begin{equation}
    \begin{aligned}
    a_t &= \pi(q,\mathcal{H}_t,\mathcal{A}_t), \\
    (\mathcal{S}_{t+1},f_t) &= \operatorname{Exec}(\mathcal{S}_t,a_t), \\
    \mathcal{H}_{t+1} &= \mathcal{H}_t\oplus(a_t,f_t).
    \end{aligned}
\end{equation}
Execution ends when the planner emits \texttt{Finish} or reaches $T_{\max}$; the assigned verifier then evaluates $\mathcal{H}_T$ and $\mathcal{S}_T$ (Figure~\ref{fig:evaluation-pipeline}).

\textbf{Stateful Mobile Environment.}
The sandbox realizes $\operatorname{Exec}$ through three layers: a \textbf{Tool Schema} presented to the planner, \textbf{Tool Implementation Code} that enforces parameter and execution rules, and a \textbf{Shared Stateful Database} that maintains application records and operational logs. We represent its state and returned feedback as
\begin{equation}
    \begin{aligned}
    \mathcal{S}_t &= \langle\mathcal{D}_t,\mathcal{O}_t\rangle, \\
    f_t &= \langle\text{Status},\text{ErrorType},\text{Payload}\rangle.
    \end{aligned}
\end{equation}
Each call validates its arguments, updates the relevant application state, records the operation, and returns a structured result or error. Resettable database snapshots and retained logs enable deterministic replay, while controlled friction such as missing arguments, permission blocks, and entity ambiguity tests whether planners can recover from execution feedback. All mobile, memory, skill-loader, and sub-agent resources follow the same function-calling protocol. Appendix~\ref{app:mobile-environment} details the action interfaces and an implementation example.

\textbf{Reference Harness.}
The reference harness assembles the request, history, and candidate schemas,
dispatches model-generated calls, and appends feedback until termination.
The model selects actions; the environment maintains state and evaluation
evidence.

\subsection{Unified Capability Coverage}
\label{sec:task-construction}
\label{sec:capability_dimensions}

We construct four task subsets to assess complementary requirements of a
central mobile planner within one executable benchmark. Each instance
specifies an initial state, candidate actions, and a capability target.
The shared environment and function-calling interface let the same planner
be evaluated across Tool Use, Memory Usage, Skill Usage, and Sub-agent
Collaboration, yielding a capability profile rather than an isolated score.
The subsets retain capability-specific requirements.

\textbf{Tool Use.}
Tool Use grounds user intent in valid tools and arguments while handling dependencies and stateful feedback. Starting from human-curated mobile requests, we introduce obfuscated references, permission constraints, intermediate state changes, and runtime errors. The tasks cover parameter grounding, dependency planning, reference and state tracking, intent revision, and boundary handling.

\textbf{Memory Usage.}
Memory Usage enables personalized execution when requests leave user-specific information implicit. We construct coherent profile worlds of identities, habits, routines, and preferences, then derive tasks that retrieve and apply those records or update persistent memory. Retrieval tasks annotate the required records and downstream objective.

\textbf{Skill Usage.}
Skill Usage evaluates reusable procedures under two action spaces. Each task has one or more acceptable skill routes, comprising one or several skills. Loading a skill exposes its instructions and bound tools. Skill-Only Routing (SOR) requires a complete route to be loaded before business-tool use, followed by task completion. Mixed Tool-Skill Routing (MTSR) also exposes direct tools and accepts valid direct-tool, skill-based, or mixed solutions; skill loading is not mandatory.

\textbf{Sub-agent Collaboration.}
Sub-agent Collaboration is required when specialized components handle GUI interaction, visual processing, conditional monitoring, or interactive practice. We construct requests around these routes and annotate the valid sub-agent and contextual handoff payload, isolating planner-side routing and handoff quality from downstream specialist execution.

\begin{figure}[!t]
    \centering
    \includegraphics[width=1.0\linewidth]{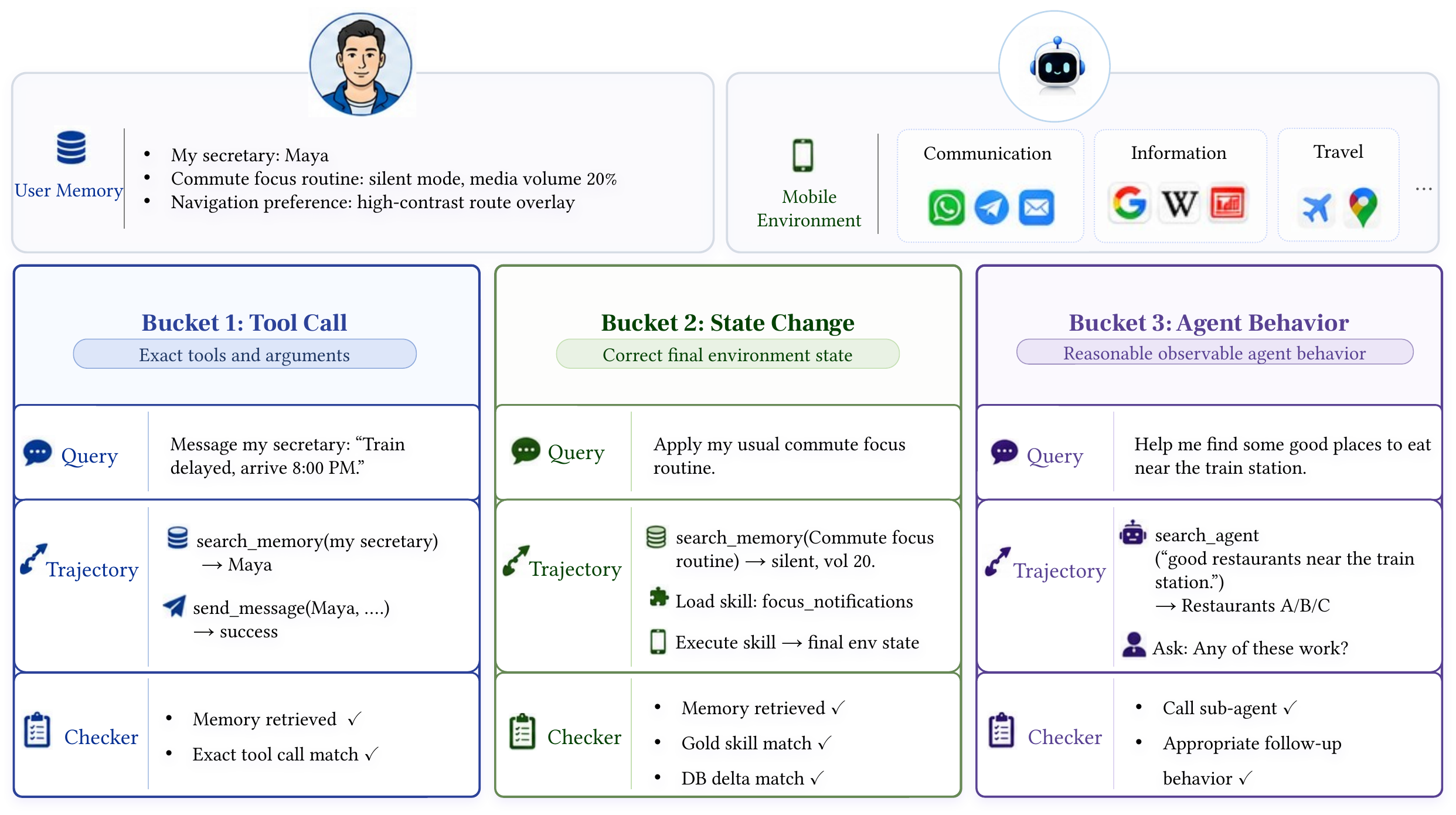}
    \caption{\textbf{Evidence-aligned verification in MobilePA-Bench.} Each query is assigned a fixed primary checker for tool calls, terminal state changes, or observable agent behavior.}
    \label{fig:evaluation-bucket}
\end{figure}

\subsection{Evidence-Aligned Verification}
\label{sec:task-evaluation}
\label{sec:metrics}

Evidence-aligned verification matches the completion criterion to the task,
rather than applying one trajectory- or state-matching rule to every query.
Exact trajectory matching can reject valid alternative plans, whereas
terminal-state checks can miss required operations, handoffs, or user-facing
interactions. Each query therefore receives a fixed primary checker during
annotation, as shown in Figure~\ref{fig:evaluation-bucket}. This assignment
determines which evidence establishes completion; the additional capability
conditions are defined in Section~\ref{sec:capability-gates}.

Let $b(q)\in\{\mathrm{tool},\mathrm{state},\mathrm{behavior}\}$ denote the bucket assigned to query $q$, and let $\mathcal{C}_{b(q)}(\mathcal{H}_T,\mathcal{S}_T)\in\{0,1\}$ denote its completion verdict, including applicable execution-validity checks. \textbf{Tool Call} checks required tools, arguments, and ordering for deterministic operations; \textbf{State Change} checks the annotated terminal database delta when multiple paths are valid; and \textbf{Agent Behavior} applies task-specific rubrics to observable routing, handoff, clarification, or recommendation behavior. The assignment is fixed before model evaluation and is not selected from the model's output. This preserves explicit operational requirements where needed while allowing terminal-state evidence for tasks with multiple valid paths. The checker outputs distinguish call-level, state-level, and interaction-level failures. 

\subsection{Capability-Gated Evaluation}
\label{sec:capability-gates}

Capability gates enforce requirements specific to each evaluation setting.
The Memory gate combines required checks, including annotated-record
retrieval when required, with task-declared action and argument constraints.
In SOR, the Skill gate requires a complete acceptable route to be loaded
before the first non-skill tool call. MTSR instead accepts any valid
completion using the available tools and skills, without a loading gate.

For a fixed setting, let $R_q\subseteq\{\mathrm{mem},\mathrm{skill}\}$ denote
the applicable gates: Skill tasks include $\mathrm{skill}$ in SOR but use
$R_q=\varnothing$ in MTSR. The task outcome is
\begin{equation}
    \operatorname{Succ}(q)
    =
    \mathcal{C}_{b(q)}(\mathcal{H}_T,\mathcal{S}_T)
    \land
    \bigwedge_{r\in R_q} g_r(q),
\end{equation}
with an empty conjunction equal to one. Skill completion includes execution
validity and both behavior judges' verdicts; MTSR rubrics accept valid
alternative paths rather than requiring the annotated skill route.
Memory addition, replacement, and removal tasks do not require retrieval
unless specified; their outcome and any applicable action constraints still
apply. Gates verify observable compliance, not causal use of retrieved
content. Route loading remains a diagnostic in MTSR, not a success
requirement. 

\subsection{Benchmark Statistics and Scoring}
\label{sec:benchmark-statistics}

MobilePA-Bench contains 1,705 distinct mobile queries and yields 1,905 scored trajectories per model: 1,040 Tool Use queries; 376 Memory queries covering retrieval-dependent execution or memory updates; 200 Skill queries evaluated under both SOR and MTSR, yielding \textbf{400 skill-evaluation instances}; and 89 Sub-agent queries covering routing, contextual handoff, and follow-up. SOR requires skill loading and completion, whereas MTSR accepts valid alternative paths. Scores assess executed task outcomes under each setting's requirements, not merely plausible final responses.
The environment provides 212 tools across 13 tool domains, distinct from the
query-level taxonomy. 

\textbf{Score Aggregation.}
\label{sec:scoring}
Overall is the weighted sum of Tool Use, Memory, Skills, and Sub-agent
scores with fixed weights of 50\%, 20\%, 20\%, and 10\%, respectively.
The authors set these weights with reference to query counts and qualitative
judgments of task complexity and difficulty, rather than a calibrated
difficulty model or measured user utility. Overall is a benchmark index;
we therefore report the full capability vector and weight-sensitivity
diagnostics alongside it.
Tool Use aggregates its task categories; Memory uses gated end-to-end
success; Skills averages the SOR and MTSR success rates, with a loading gate
only in SOR; and Sub-agent measures routing-and-handoff success. All
scores retain the full denominator, counting missing or invalid predictions
as failures.

\section{Experiments}
\label{sec:experiments}

We first examine the capability profiles exposed by unified coverage, then
use the recorded checks to separate capability-gate failures from downstream
completion failures. Reliability analyses assess score variation and
weight sensitivity. Finally, small-scale online RL examines the sandbox's
utility for planner development.

\subsection{Evaluation Protocol}
\label{sec:exp_setup}

We evaluate 14 frontier LLMs on all 1,705 MobilePA-Bench tasks. Because each of
the 200 Skill Usage tasks is executed under both Skill-Only Routing (SOR) and
Mixed Tool-Skill Routing (MTSR), the evaluation contains 1,905 scored
trajectories per model. Candidate lists are supplied with the tasks, rather
than retrieved online by the evaluated model. Non-Skill tasks expose 15 tools.
Skill evaluation uses 20 candidate skills in SOR and 15 direct tools plus
five candidate skills in MTSR; loading a skill exposes its bound tools.
Thus, this comparison changes both direct-tool access and the skill
candidate set, rather than isolating the effect of direct tools.
Every model uses the reference harness and the same
function-calling interface until emitting \texttt{Finish} or
reaching $T_{\max}=15$. Missing or invalid predictions count as failures. Each
task uses its pre-assigned primary checker and any required capability gate;
API-judged outcomes use a fixed unanimous-AND decision from two judges.
Appendix~\ref{app:evaluation-protocol} reports reference-tool accessibility
and provenance checks; these do not establish open-ended retrieval quality.

\subsection{Overall Performance}
\label{sec:main_results}

Table~\ref{tab:main_results} reports the recorded capability and Overall
scores under the setting-specific success criteria and fixed aggregation
defined in Sections~\ref{sec:capability-gates}--\ref{sec:benchmark-statistics}.

\begin{table}[!t]
\centering

\small
\setlength{\tabcolsep}{6pt}
\begin{tabular*}{\textwidth}{@{\extracolsep{\fill}}lcccccc@{}}
\toprule
\textbf{Model} & \textbf{Size / Access} & \textbf{Overall} & \textbf{Tool Use} & \textbf{Memory} & \textbf{Skills} & \textbf{Sub-agent} \\
\midrule
GPT-6-Astra & Closed source & \textbf{76.87} & 75.77 & 74.73 & \textbf{93.25} & 53.93 \\
Claude-Opus-5 & Closed source & 75.71 & \textbf{77.60} & 71.81 & 83.00 & 59.55 \\
Claude-Fable-5 & Closed source & 74.53 & 77.21 & \textbf{76.33} & 69.00 & \textbf{68.54} \\
Gemini-3.6-Flash & Closed source & 69.61 & 74.71 & 69.68 & 65.75 & 51.69 \\
Gemini-3.1-Pro & Closed source & 68.16 & 75.38 & 65.69 & 55.75 & 61.80 \\
Claude-Opus-4.8 & Closed source & 66.05 & 72.40 & 56.38 & 69.25 & 47.19 \\
Qwen-3.7-Max & Closed source & 64.89 & 72.31 & 64.10 & 53.75 & 51.69 \\
Seed-2.1-Pro & Closed source & 64.89 & 70.29 & 57.18 & 64.00 & 55.06 \\
GLM-5.3 & 744B-A40B & 73.88 & 76.44 & 72.07 & 77.00 & 58.43 \\
Qwen-3.8-Max & 2.4T-A95B & 71.77 & 73.65 & 73.14 & 75.75 & 51.69 \\
Kimi-K3 & 2.8T-A104B & 69.64 & 71.54 & 71.01 & 74.75 & 47.19 \\
GLM-5.3-Flash & 320B-A18B & 68.31 & 71.35 & 65.16 & 68.25 & 59.55 \\
GLM-5.2 & 744B-A40B & 66.06 & 73.75 & 60.37 & 58.00 & 55.06 \\
\bottomrule
\end{tabular*}
\caption{\textbf{MobilePA-Bench results (\%).} Overall uses the fixed
50/20/20/10 weighting in Section~\ref{sec:scoring}. Bold marks the highest
observed score in each column. Sizes use B (billion), T (trillion), and
A (active parameters). Full-cohort analyses include GPT-5.6-Sol
(Appendix~\ref{app:capability-diagnostics}).}
\label{tab:main_results}
\end{table}

\begin{figure}[!t]
    \centering
    \includegraphics[width=\linewidth]{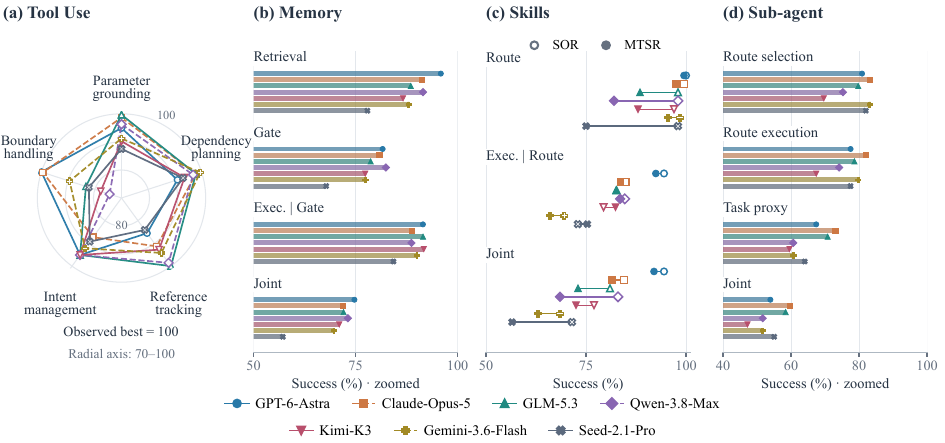}
    \caption{\textbf{Fine-grained capability profiles of seven family representatives}
    (highest Overall per family; Table~\ref{tab:main_results}).
    (a) Tool Use normalized to each subtype's best observed score across
    14 models ($=100$). (b--d) Rates (\%); (c) Skill route diagnostics
    (SOR: open markers; MTSR: filled). }
    \label{fig:capability-diagnosis}
\end{figure}

\texttt{GPT-6-Astra} has the highest Overall point estimate at $76.87\%$,
while nine of the 14 models score below $70\%$. Its observed lead over
\texttt{Claude-Opus-5} is $1.16$ percentage points. The dimension leaders differ:
\texttt{Claude-Opus-5} leads Tool Use, \texttt{Claude-Fable-5} leads Memory
Usage and Sub-agent Collaboration, and \texttt{GPT-6-Astra} leads Skill Usage.

Across models, the mean scores are $73.31\%$ for Tool Use, $66.13\%$ for
Memory Usage, $70.14\%$ for Skill Usage, and $54.18\%$ for Sub-agent
Collaboration. Skill Usage has the widest range ($39.50$ points), whereas no
model leads all four dimensions. Unified coverage therefore exposes planner
profiles that an Overall score alone would hide. These are differences on
the benchmark's task subsets, not a difficulty-calibrated ranking of the
underlying capabilities.

\subsection{Fine-Grained Capability Diagnosis}
\label{sec:diagnostic-analysis}

Beyond aggregate scores, MobilePA-Bench examines four complementary
capabilities through behavioral categories and execution-level checks.
Figure~\ref{fig:capability-diagnosis} presents seven model-family
representatives, revealing both uneven capability profiles and differences
between resource access, prescribed behavior, and task completion. This
decomposition identifies weaknesses that aggregate rankings can obscure.
Full-cohort summaries and descriptive Skill comparisons are provided in
Appendix~\ref{app:capability-diagnostics}.

\textbf{Tool Use: Uneven behavioral profiles.}
Strong performance in one type of tool use does not imply consistent
performance across others. Figure~\ref{fig:capability-diagnosis}(a)
visualizes these differences relative to the highest observed score in each
subtype. In raw accuracy, \texttt{GLM-5.3} and \texttt{Qwen-3.8-Max}
achieve $94.5\%$ and $93.2\%$ on Reference Tracking, but only $67.0\%$ and
$59.8\%$ on Boundary Handling. Parameter Grounding and Dependency Planning
also remain below $80\%$ for every selected representative. These results
distinguish proficiency in tracking referenced entities from reliability in
grounding arguments, satisfying prerequisites, and handling execution
boundaries, providing more specific improvement targets than a single Tool
Use score.

\textbf{Memory: Distinguishing unmet requirements from completion failures.}
Retrieving the required records is only one component of Memory
performance. Figure~\ref{fig:capability-diagnosis}(b) separates retrieval,
compliance with the complete memory gate, and task completion. Across all
seven representatives, more tasks fail the complete gate than pass it but
fail completion. For \texttt{Qwen-3.8-Max}, these two groups contain 66 and
35 of the same 376 tasks, respectively. This decomposition distinguishes
unmet retrieval or action requirements from completion failures among
gate-passing trajectories, directing attention to different aspects of
memory-supported execution. Retrieval and conditional execution use
different task subsets, so the four displayed rates should not be
interpreted as a sequential attrition funnel.

\textbf{Skills: Separating skill routing from task execution.}
Correct skill loading does not guarantee successful execution. In SOR, the
seven representatives achieve Route rates of $97.0\%$--$100.0\%$, while
Joint success ranges from $68.5\%$ to $94.5\%$
(Figure~\ref{fig:capability-diagnosis}(c)). \texttt{Gemini-3.6-Flash}, for
example, achieves $98.5\%$ Route but only $68.5\%$ Joint success, exposing
substantial completion failures despite satisfying the routing requirement.
The mixed configuration reveals further differences in routing behavior:
\texttt{Qwen-3.8-Max}'s ordered-route pass rate changes from $98.0\%$ in
SOR to $82.0\%$ in MTSR. However, a lower reference-route pass rate does not
establish lower task completion. MTSR accepts valid direct-tool, skill-based,
and mixed solutions; Route and Joint diagnose reference-route use, not
MTSR task success. Because both
tool access and skill candidates differ between settings, this comparison
characterizes task-dependent routing rather than isolating
direct-tool interference.

\Needspace{4\baselineskip}
\textbf{Sub-agent: Correct routing is insufficient for complete interaction.}
Figure~\ref{fig:capability-diagnosis}(d) distinguishes selecting a
specialist, executing the routing action, and satisfying the complete
task-level interaction requirements. \texttt{Claude-Opus-5} and
\texttt{Gemini-3.6-Flash} both achieve $83.1\%$ route selection, yet their
Joint scores are $59.6\%$ and $51.7\%$, respectively. Thus, similar routing
accuracy can conceal different levels of complete interaction success.
These checks motivate closer examination of handoff and follow-up behavior
rather than treating a correct invocation as successful collaboration. The
results concern planner-side interactions over 89 tasks, not downstream
specialist execution; the saved task-success proxy is also not an
independent measure of handoff quality.

Together, these analyses combine coverage across four capabilities with
finer distinctions within each capability. They move evaluation beyond
identifying which model scores higher toward identifying which observable
behaviors warrant further investigation. The resulting profiles can help
prioritize targeted training and validation for argument grounding,
dependency handling, memory requirements, skill-supported execution, and
delegation behavior, without assuming that the observed differences
establish their internal causal mechanisms.

\subsection{Reliability and Score Sensitivity}
\label{sec:reliability-failures}

\textbf{Evaluation Stability.}
Three complete evaluations of an additional \texttt{Qwen3.8-27B}
configuration yield Overall scores of $66.60\%$--$67.35\%$ (sample standard
deviation $0.38$ percentage points), with Memory the most variable dimension
(standard deviation $2.15$ points). This model is distinct from
\texttt{Qwen-3.8-Max} in Table~\ref{tab:main_results}; these runs do not
establish the rerun variability of all 14 models.

\textbf{Ranking Sensitivity.}
We compare the default weighting with equal-dimension and equal-evaluation-instance weighting. Both preserve the top-six order. Equal-dimension weighting shifts five models by at most two positions; instance weighting only swaps ranks 10 and 11. The leading order is stable under these alternatives, but Overall should be read alongside capability-specific scores.

\subsection{Online RL Baseline Model}
\label{sec:online-rl}

We initialize the baseline from \texttt{Qwen3.8-27B} and construct 4,000
training queries disjoint from the benchmark's test queries. We train in
the simulated mobile sandbox using Group Relative Policy Optimization
(GRPO), with executable tools and structured
environment feedback. Rewards are supplied by a learned judge model trained
on trajectories collected during preliminary evaluations of API-based
models. Table~\ref{tab:online-rl} compares the trained checkpoint with the
original model's mean over three complete evaluations under the same
benchmark protocol.

Overall improves from $66.94\%$ to $69.25\%$, a gain of $2.31$ percentage
points. Tool Use, Memory Usage, Skill Usage, and Sub-agent Collaboration
increase by $3.06$, $0.48$, $2.33$, and $2.25$ points, respectively.
Tool Use improves most, while the smaller Memory gain indicates uneven
progress across capabilities. These results illustrate the benchmark's
dual role in supporting online planner optimization and locating
capability-specific improvements. This is a single-checkpoint comparison,
not evidence from repeated independent training runs.

\input{src/tables/exploratory_rl}

\section{Conclusion}
\label{sec:conclusion}

We introduced \textbf{MobilePA-Bench} to evaluate central mobile planners
across Tool Use, Memory Usage, Skill Usage, and Sub-agent Collaboration
in a shared stateful environment. Evidence-aligned checkers and
setting-specific capability gates connect task outcomes with resource
and procedural requirements, supporting diagnosis beyond aggregate success.
The experiments show that retrieving memories, loading skills, or selecting
a sub-agent does not by itself satisfy the full task requirements. An online-RL
baseline also improves all four capability scores under the reported
protocol, providing initial evidence that the sandbox can support planner
optimization as well as evaluation. Together, these results establish a
testbed for identifying mobile planning weaknesses and assessing targeted
improvements.

\label{sec:main-text-end}
\section*{Limitations}
\label{sec:limitations}
MobilePA-Bench evaluates central planning through supplied tool interfaces
in a simulated environment. It does not measure visual grounding,
low-level GUI execution, or real-device variability. Sub-agent evaluation
covers planner-side routing and handoff rather than downstream specialist
performance; transfer to real-device workflows remains unverified.
The four capabilities are evaluated on distinct task subsets, rather than
jointly in every trajectory. Coverage is finite, especially for the
89 Sub-agent tasks; difficulty is not calibrated across subsets. Gates
verify observable compliance, not causal use of memory or skill contents.
Overall is a benchmark index, not a user-experience ranking. 

\bibliographystyle{unsrtnat}
\bibliography{references}

\clearpage
\appendix
\section*{Appendix}
\input{appendix}

\end{document}

%% file: authors.tex
\author[1]{
{\small\textnormal{Yi Zhu$^{*}$, Xiongwei Wu$^{*}$, Qiyi Wang$^{\ddagger}$, Tingyu Qu, Jiajun Liu, Sihan Cao,}\\
\vspace{-0.6em}
\small\textnormal{Long Chen, Weigao Sun, Feida Zhu$^{\S}$, Yiran Zhong$^{\S}$, Steven HOI\\}}
\normalsize
\vspace{1em}
\textbf{MAI Team\UniMobileBenchLogo , Alibaba Token Hub, Alibaba Group} \\
\vspace{0.3em}
{\footnotesize\faGithub\enspace\textnormal\url{https://github.com/Tongyi-MAI/MobilePA-Bench}}
\vspace{-1cm}
}

%% file: src/tables/exploratory_rl.tex
\begin{table}[H]
\centering
\footnotesize
\setlength{\tabcolsep}{2pt}
\renewcommand{\arraystretch}{1.10}
\fitwide{%
\begin{tabular}{@{}lccccc@{}}
\toprule
\makecell[l]{\textbf{Model /}\\\textbf{Setting}} & \textbf{Overall} & \makecell{\textbf{Tool}\\\textbf{Use}} &
\textbf{Memory} & \textbf{Skills} & \makecell{\textbf{Sub-}\\\textbf{agent}} \\
\midrule
\makecell[l]{Qwen3.8-27B\\(mean)} & 66.94 & 68.36 & 67.64 & 72.92 & 46.44 \\
\rowcolor{gray!10}
\quad + Online RL & \textbf{69.25} & 71.42 & 68.12 & 75.25 & 48.69 \\
$\Delta$ (points) & $+2.31$ & $+3.06$ & $+0.48$ & $+2.33$ & $+2.25$ \\
\bottomrule
\end{tabular}}
\caption{\textbf{Small-scale online RL in the MobilePA-Bench sandbox (\%).}
Skills/Overall retain legacy gating. Baseline: three-run mean;
$\Delta$: improvement over that mean in percentage points.}
\label{tab:online-rl}
\end{table}

%% file: appendix.tex
\raggedbottom

\section{Benchmark and Tool Statistics}
\label{app:benchmark-statistics}

We first summarize the query set and tool inventory. Query domains describe
user tasks, whereas tool domains organize the environment's executable APIs;
the two taxonomies serve different purposes.

\subsection{Dataset Composition and Query Coverage}
\label{app:query-statistics}

MobilePA-Bench contains 1,705 distinct queries: 1,040 for Tool Use,
376 for Memory Usage, 200 for Skill Usage, and 89 for Sub-agent Collaboration.
The same 200 Skill queries are evaluated under both SOR and MTSR,
resulting in 1,905 evaluation instances per model rather than additional
distinct queries.

\begin{figure}[H]
    \centering
    \includegraphics[width=0.94\textwidth]{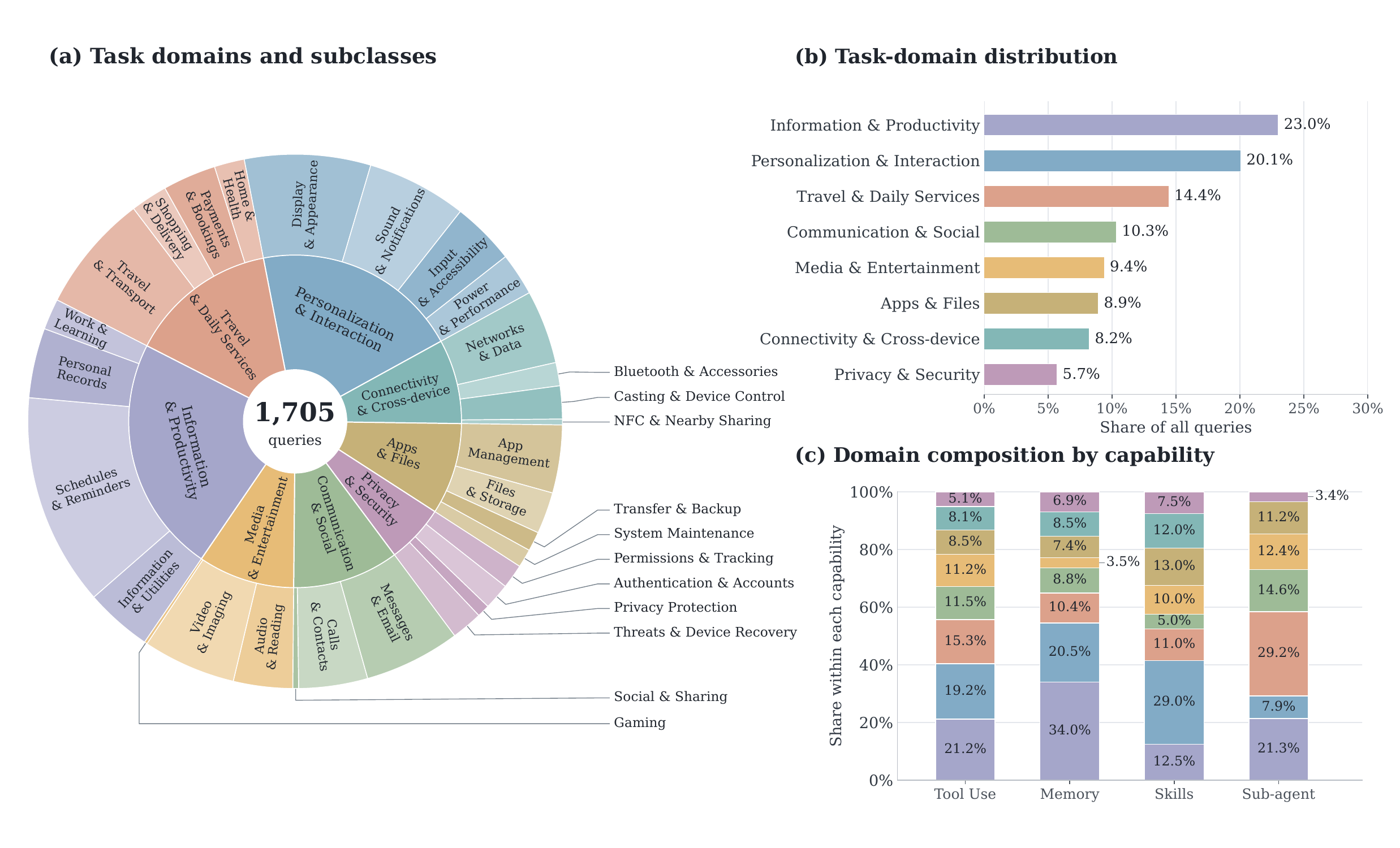}
    \caption{\textbf{Descriptive coverage of 1,705 queries.}
    (a) Query domains and subclasses; (b) domain shares;
    (c) domain composition within each capability, normalized to 100\%.
    Colors are consistent across panels; stacks in (c) follow (b) from bottom to top.}
    \label{fig:benchmark-statistics}
\end{figure}

Figure~\ref{fig:benchmark-statistics} summarizes all 1,705 tasks under an
AI-assisted descriptive taxonomy of eight query domains and 30 subclasses.
This coverage taxonomy is descriptive and does not define
the four official capability scores.

\subsection{Memory Task Statistics}
\label{app:memory-statistics}

\begin{figure}[!t]
    \centering
    \includegraphics[width=0.75\textwidth]{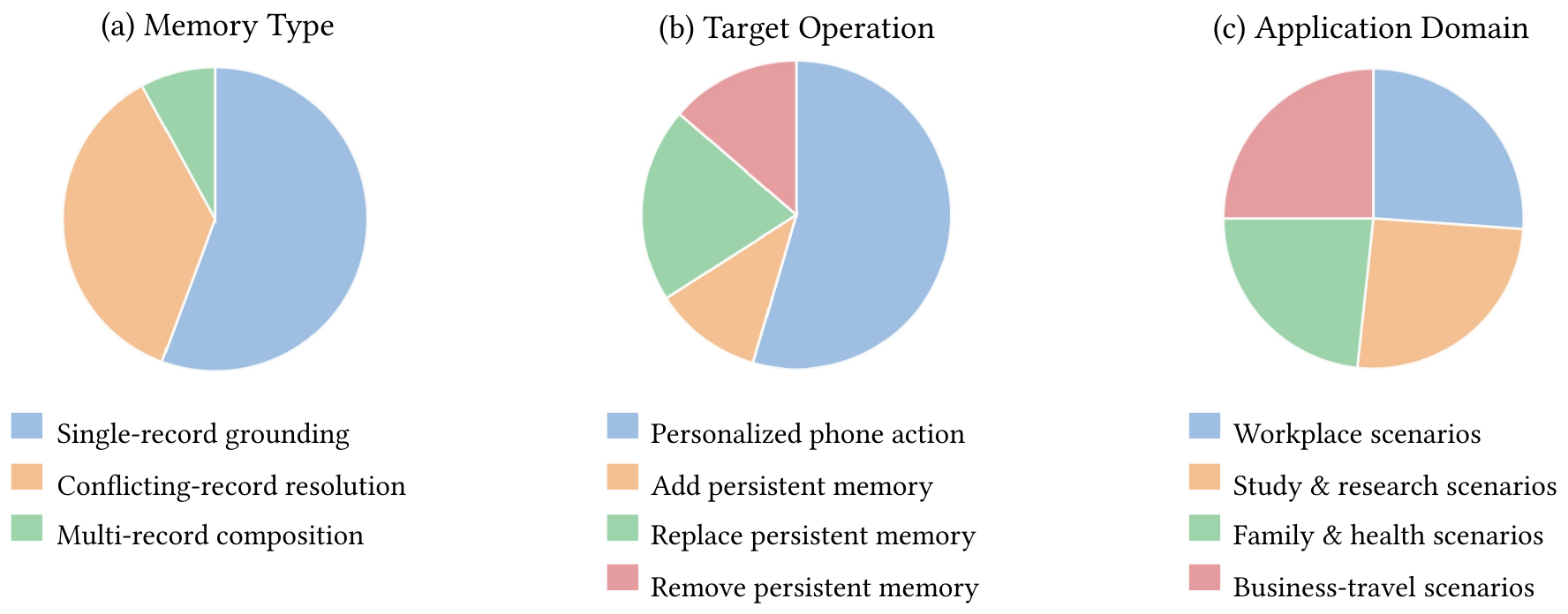}
    \caption{\textbf{Coverage of the 376 Memory Usage tasks along three diagnostic axes:} (a) memory reasoning type (single-record grounding, conflicting-record resolution, and multi-record composition); (b) target operation (personalized phone actions and memory addition, replacement, or removal); and (c) application domain.}
    \label{fig:memory_coverage}
\end{figure}

Figure~\ref{fig:memory_coverage} characterizes the 376 Memory Usage tasks
by memory reasoning type, target operation, and application domain.
These complementary views describe the information a task requires,
the operation it targets, and its application context.

\subsection{Tool Inventory by Domain}
\label{app:tool-statistics}

Table~\ref{tab:tool_domains} lists the 212 tools across 13 tool domains.
This inventory describes the available action interfaces, not the
query-domain distribution in Figure~\ref{fig:benchmark-statistics}.

\begin{table}[!t]
\centering

\small
\setlength{\tabcolsep}{5pt}
\renewcommand{\arraystretch}{1.1}
\begin{tabularx}{\linewidth}{l r >{\scriptsize\justifying\arraybackslash}X}
\toprule
\textbf{Domain} & \textbf{\# Tools} & \textbf{Description} \\
\midrule
Audio \& Entertainment & 25 & Media control, playback, camera, screen recording, and music recognition. \\
\addlinespace[0.2em]
Apps \& Storage & 23 & App lifecycle, installations, permissions, notifications, and storage cleanup. \\
\addlinespace[0.2em]
Display \& Sound & 22 & Brightness, volume, sound modes, dark mode, DND, and power schedules. \\
\addlinespace[0.2em]
System Settings & 22 & Control center, device info, desktop layout, language, and system updates. \\
\addlinespace[0.2em]
Time Management & 16 & Calendar schedules, tasks, alarms, timers, and automation rules. \\
\addlinespace[0.2em]
AI Assistant & 16 & GUI sub-agent routing, image processing, memory search, and smart perception. \\
\addlinespace[0.2em]
Calls \& Communication & 15 & Contacts, calls, SMS/email handling, and harassment interception. \\
\addlinespace[0.2em]
Network \& Connectivity & 14 & WLAN, Bluetooth, mobile data, hotspot, NFC, and pairing management. \\
\addlinespace[0.2em]
Travel \& Lifestyle & 13 & Weather, navigation, transit tracking, payment codes, and ticket booking. \\
\addlinespace[0.2em]
Devices \& Cross-device & 13 & Screen casting, multi-screen sharing, device migration, and pairing logs. \\
\addlinespace[0.2em]
Input \& Interaction & 12 & Screenshots, QR scanning, air gestures, system buttons, and AOD modes. \\
\addlinespace[0.2em]
Utilities \& Productivity & 11 & Browser, calculator, conversions, translations, and open-domain QA. \\
\addlinespace[0.2em]
Security \& Privacy & 10 & Screen lock, app locks, biometrics, password vault, and emergency alerts. \\
\bottomrule
\end{tabularx}
\caption{Tool domains in the MobilePA-Bench mobile function-call environment.}
\label{tab:tool_domains}
\end{table}

\section{Extended Experimental Results}
\label{app:extended-experiments}

\subsection{Evaluation Configuration and Provenance}
\label{app:evaluation-protocol}

The reported results use matched task definitions and a common aggregation
policy for all 14 models. Each model
has 1,040 Tool Use trajectories, 376 Memory Usage trajectories, 89 Sub-agent
Collaboration trajectories, and 400 Skill Usage trajectories obtained by
executing the same 200 tasks under SOR and MTSR. Thus, the 1,705 distinct tasks
produce 1,905 scored trajectories per model. API-judged outcomes use the same
two-judge unanimous-AND policy throughout; saved database results and required
memory or skill gates are retained where applicable.

The results follow the setting-specific Skill criteria in
Appendix~\ref{app:verifier-details}: SOR requires an acceptable skill route
and task completion, whereas MTSR accepts valid direct-tool, skill-based,
or mixed completions without a loading gate. Reported Skills and Overall
scores, ranking comparisons, and online-RL results use these criteria.
Route-conditioned diagnostics are separate from MTSR task success.

Before aggregation, we checked row counts, task identifiers, query text, and
evaluation-policy fields across the canonical result manifests. These checks
establish matched benchmark content and scoring policy, but do not by
themselves prove byte-identical historical tool implementations, initial-state
serialization, or checker code. Missing or invalid predictions remain in the
full denominator as failures.

\paragraph{Candidate actions.}
Candidate schemas are supplied with each task, not produced by a learned
online retriever in this evaluation. Non-Skill inputs expose 15 tools per
task. SOR exposes 20 skills without direct business tools; MTSR exposes
15 direct tools and five skills. All 14 models' recorded Skill candidate
lists confirm these sizes and contain at least one complete annotated route
for each task. Loading a skill adds its bound tools to the active action set.
These configurations differ in both tool access and skill candidates,
so their comparison is not a single-factor direct-tool ablation.

An offline audit of the task inputs found that every tool named in nonempty
reference calls is present among the supplied candidates: 841/841 Tool Use
tasks, 89/89 Sub-agent tasks, and 176/176 single-turn plus 200/200 multi-turn
Memory tasks. The remaining 199 Tool Use tasks have empty reference-call
lists and are not counted as successful recall cases. Skill inputs do not
provide an independent reference-call list, so we report route availability
rather than business-tool recall for them. These checks establish access to
the recorded reference tools or routes, not the necessity of those references,
coverage of all valid solutions, or general retrieval quality. Historical
Skill lists are checked against saved trajectories; input-level coverage
does not establish byte-identical runtime schemas for every historical run.

\paragraph{Model identities and reproducibility scope.}
The additional \texttt{Qwen3.8-27B} model used for stability and exploratory
training is not the \texttt{Qwen-3.8-Max} entry in the main comparison.
Historical task-ID and policy matching do not establish access to identical
future API snapshots. The main results retain the recorded judgments and
execution outcomes; the main-text weighting comparisons hold these records fixed.

\subsection{Full-Cohort Capability Summaries}
\label{app:capability-diagnostics}

\begin{figure}[!t]
    \centering
    \includegraphics[width=\linewidth]{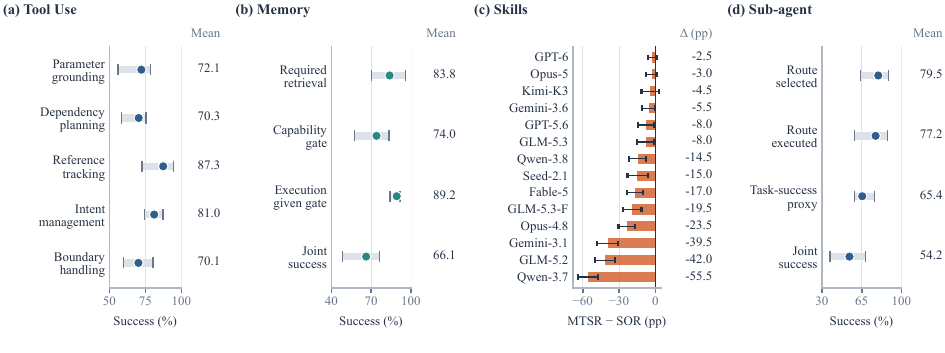}
    \caption{\textbf{Capability diagnosis across 14 models.}
    (a,b,d): unweighted means and model ranges (not confidence intervals);
    axes start above zero. (c): paired Skill Joint-diagnostic differences
    (MTSR--SOR), shown as point estimates. The Sub-agent task-success
    proxy is not an independent handoff-quality measure.}
    \label{fig:capability-diagnosis-full}
\end{figure}

Figure~\ref{fig:capability-diagnosis-full} complements the seven-model
profiles in Figure~\ref{fig:capability-diagnosis} with aggregate diagnostics
for all 14 models. Panels (a), (b), and (d) summarize means and model ranges;
panel (c) reports paired MTSR--SOR differences in the Skill Joint diagnostic as
point estimates. These summaries retain the full cohort
rather than selecting one representative per model family.

Across the full cohort, mean Tool Use accuracy is $87.3\%$ for Reference
Tracking and $81.0\%$ for Intent Management, compared with $72.1\%$ for
Parameter Grounding, $70.3\%$ for Dependency Planning, and $70.1\%$ for
Boundary Handling. Memory's mean Retrieval, Gate, conditional completion,
and Joint rates are $83.8\%$, $74.0\%$, $89.2\%$, and $66.1\%$; their
different denominators preclude interpreting them as one attrition funnel.
Every model's Skill Joint diagnostic
is lower in MTSR than in SOR. The largest
decreases occur for \texttt{Qwen-3.7-Max} ($55.5$ percentage points),
\texttt{GLM-5.2} ($42.0$), and \texttt{Gemini-3.1-Pro} ($39.5$). These
route-conditioned differences do not measure differences in setting-specific
task success.
Mean Sub-agent Route, route execution, Task Proxy, and Joint rates are
$79.5\%$, $77.2\%$, $65.4\%$, and $54.2\%$, respectively. These summaries
describe the fixed 14-model cohort, not confidence intervals over models.

\subsection{Exploratory Online RL}
\label{app:online-rl}

An online-RL checkpoint initialized from \texttt{Qwen3.8-27B} was trained
through the simulated mobile environment using 4,000 training queries
disjoint from the benchmark's test queries. Table~\ref{tab:online-rl}
compares its scores with the original model's mean over three complete
evaluations. The reported Overall improvement is $2.31$ percentage points.
Query-level disjointness does not establish separation of templates, profile
worlds, or tool schemas. The comparison uses one trained checkpoint rather
than repeated independent training runs and therefore remains exploratory.

\section{Mobile Environment Details}
\label{app:mobile-environment}

This appendix provides the action-interface definitions and sandbox example
supporting Section~\ref{sec:sandbox}. The tool inventory is reported in
Appendix~\ref{app:tool-statistics}.

\subsection{Unified Action Interfaces}

All planner resources are represented as structured function schemas with a unique name, textual description, and JSON argument schema:
\begin{itemize}
    \item \textbf{Direct Mobile Tools} query or mutate mobile-system state across domains such as messaging, settings, media, and calendars.
    \item \textbf{Memory Tools} search persistent user-profile and contextual records and return ranked results as structured feedback.
    \item \textbf{Skill Loading Tools} return procedural instructions and expose the concrete tools associated with a reusable skill.
    \item \textbf{Sub-agent Entry Tools} route a task and its contextual payload to a specialized downstream agent.
\end{itemize}

Let $\mathcal{G}$ denote the catalog of direct, memory, and sub-agent tools,
and let $\mathcal{L}_q$ denote the candidate skill loaders for query $q$.
The supplied task configuration defines a candidate tool set
$\mathcal{C}_q\subseteq\mathcal{G}$; the initial action set is
\begin{equation}
    \mathcal{A}_0=\mathcal{C}_q\cup\mathcal{L}_q.
\end{equation}
This notation does not assume a learned online retriever. In SOR,
$\mathcal{C}_q$ is empty and $|\mathcal{L}_q|=20$; in MTSR,
$|\mathcal{C}_q|=15$ and $|\mathcal{L}_q|=5$.
Loading skill $s$ expands the next-step action set with the tools bound to that skill:
\begin{equation}
    \mathcal{A}_{t+1}=\mathcal{A}_t\cup\mathcal{G}(s),
\end{equation}
while the action set remains unchanged when no skill is loaded.

\subsection{Sandbox Execution Example}

\begin{figure}[!t]
    \centering
    \includegraphics[width=0.72\linewidth]{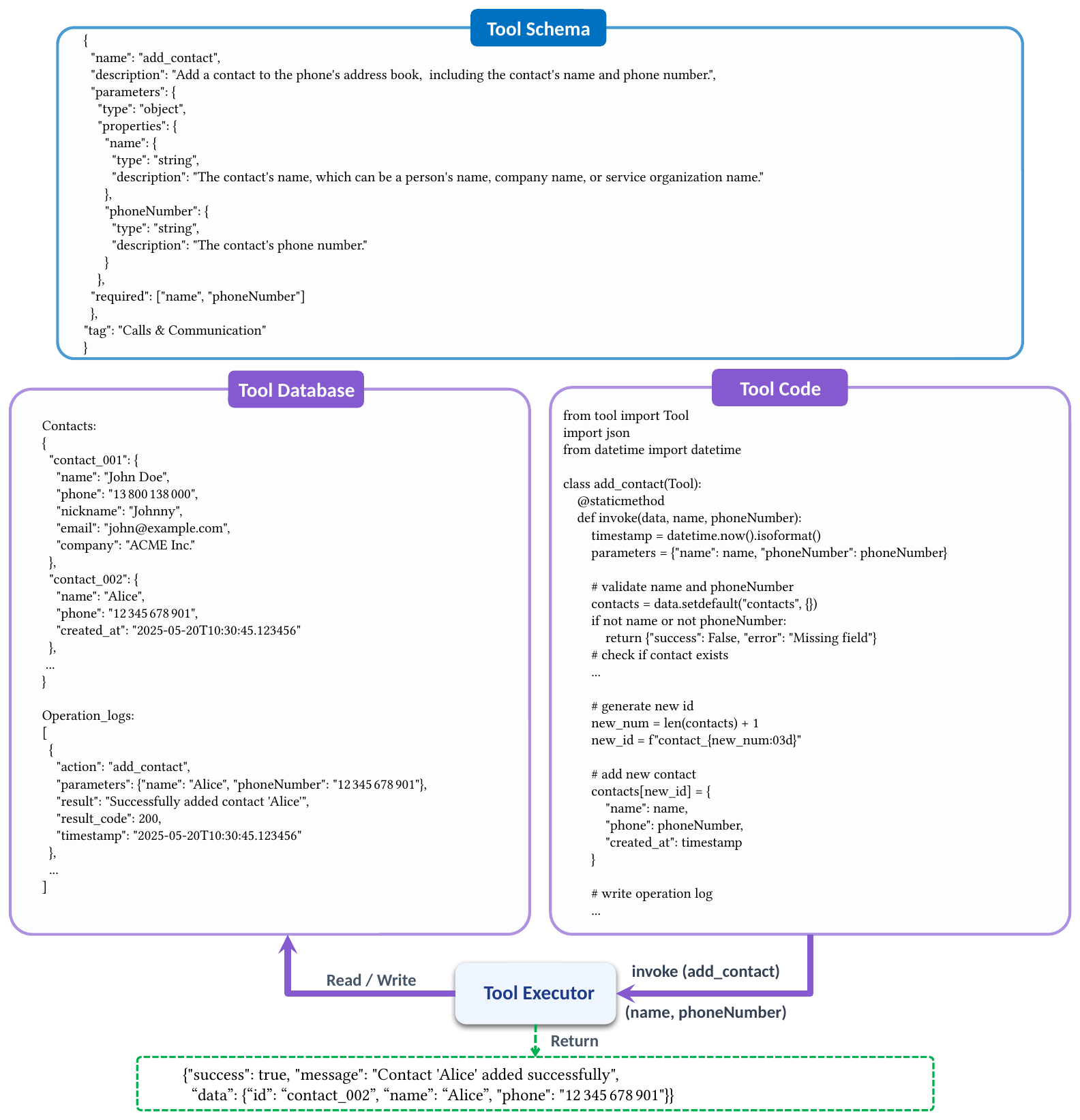}
    \caption{\textbf{Stateful sandbox architecture for an \texttt{add\_contact} invocation.} A schema defines the interface, implementation code validates and executes the call, and a shared database records state changes and operational logs. The executor returns structured feedback to the planner.}
    \label{fig:sandbox_architecture}
\end{figure}

Figure~\ref{fig:sandbox_architecture} illustrates how the shared schema, implementation, and state layers instantiate an \texttt{add\_\allowbreak contact} call.

\section{Capability-Specific Task Construction}
\label{app:task-coverage}

Section~\ref{sec:task-construction} describes the motivation and construction principle for each capability. Here we provide the detailed behavioral categories, action-space settings, and annotation scope used to instantiate those tasks.

\paragraph{Tool Use.}
Tool Use tasks are synthesized from human-curated mobile seeds and introduce realistic execution friction through reference obfuscation, permission blocks, missing arguments, and intermediate state changes. The five behavioral categories are:
\begin{itemize}
    \item \textbf{Tool and Parameter Grounding}: selecting the target API and resolving explicit or implicit arguments;
    \item \textbf{Conditional and Dependency Planning}: satisfying prerequisite order and preserving cross-step dependencies;
    \item \textbf{Reference and State Tracking}: resolving historical entities, references, and evolving application state;
    \item \textbf{Intent Revision and Task Management}: handling corrections, target changes, multi-intent requests, and continuation; and
    \item \textbf{Boundary Detection and Error Recovery}: recognizing capability limits and recovering from contradictory constraints or runtime errors.
\end{itemize}
These categories are mutually exclusive and contain 377, 249, 73, 117, and
224 tasks, respectively.
Deterministic operations use Tool Call verification, whereas path-equivalent workflows use State Change verification.

\paragraph{Memory Usage.}
Memory tasks are derived from coherent profile worlds containing user habits, identities, routines, and historical preferences. Requests intentionally omit relevant personal information, and each example records the memory IDs required to resolve the request together with its downstream execution objective. Appendix~\ref{app:memory-statistics} characterizes these tasks by memory reasoning type, target operation, and application domain.

\paragraph{Skill Usage.}
Composite mobile routines are organized into reusable skill packages. Each task specifies an execution objective and one or more acceptable skill routes. In \textit{Skill-Only Routing (SOR)}, skill loaders are initially available without direct business tools; all skills in at least one acceptable route must be loaded before business-tool use, and the task must then be completed. \textit{Mixed Tool-Skill Routing (MTSR)} exposes candidate skill loaders together with supplied direct tools. It accepts valid direct-tool, skill-based, or mixed solutions within the available action space, without requiring an annotated route. Reference routes support SOR verification and route diagnostics in both settings, rather than defining a unique solution for MTSR.

\paragraph{Sub-agent Collaboration.}
Sub-agent tasks cover specialized routes such as GUI manipulation, visual processing, conditional monitoring, and interactive practice. Annotations specify the valid downstream route and the contextual payload required for handoff. Evaluation stops at routing and handoff quality and does not score the downstream sub-agent policy.

\section{Verifier and Capability-Gate Details}
\label{app:verifier-details}

\subsection{Primary Checkers}

Each query receives exactly one primary checker before model evaluation.

\paragraph{Tool Call.}
This checker is used when the annotated solution requires a deterministic operational sequence. It verifies required tool names, call order, argument fields, and normalized argument values, while rejecting extraneous calls with task-relevant side effects.

\paragraph{State Change.}
This checker is used when multiple action sequences may reach the same valid outcome. It compares the terminal database transition $\mathcal{D}_T-\mathcal{D}_0$ with the annotated target delta and rejects destructive or unrelated writes.

\paragraph{Agent Behavior.}
This checker is used when neither an exact call sequence nor a static database mutation fully captures completion. Sub-agent tasks use two explicit rubric fields: \textit{Call Sub-agent} checks route selection, and \textit{Appropriate Follow-up Behavior} checks the handoff payload and subsequent response. Other task-specific rubrics inspect necessary clarification or presentation of a user-facing result.

\subsection{Memory and Skill Gates}
\label{app:memory-skill-gates}

\paragraph{Memory retrieval and action compliance.}
Let $\mathcal{M}_q^*$ denote the annotated memory IDs required by query $q$.
The set returned by memory search over the trajectory is
\begin{equation}
    \widehat{\mathcal{M}}_T
    =
    \bigcup_{\substack{t:\,a_t=\\\texttt{search\_user\_memory}}}
    \operatorname{MemoryIDs}(f_t).
\end{equation}
On retrieval-required tasks, the retrieval check passes only if all required
IDs occur in the recorded search results and those results are valid.
Partial retrieval is not a task-level pass. The reported Retrieval rate uses
the 367 tasks requiring this check, not all 376 Memory tasks.

The complete Memory gate additionally includes the task's declared action
constraints. Let $\mathcal{Q}_q$ contain the required checker predicates,
including the retrieval check when applicable, and let $a_q$ denote the
action-policy verdict. Then
\begin{equation}
    g_{\mathrm{mem}}(q)
    =
    a_q \land \bigwedge_{h\in\mathcal{Q}_q}h(q).
\end{equation}
An empty conjunction and an absent action policy each evaluate to one;
required checks with missing or invalid evidence fail. Where specified,
the action policy checks required actions and arguments, forbids
unauthorized state-changing calls, and applies the task-declared allowances
for read-only helpers or recoverable attempts. These checks supplement,
rather than replace, database-state or behavior-based completion checks.

Memory addition, replacement, and removal tasks without a retrieval
requirement are evaluated by their fixed primary checker and any applicable
action constraints. Gate and Joint rates use all 376 tasks.
The retrieval check scans the whole trajectory and does not impose
retrieval-before-action ordering. It establishes observable access to the
required records, not causal use of their contents.

\paragraph{Memory metric definitions.}
For each model, let $N_R=367$ be the number of retrieval-required tasks
and $n_R$ the number passing the retrieval check. Across all $N=376$
Memory tasks, let $n_G$ count complete-gate passes and $n_{G\cap T}$
count tasks passing both the gate and the primary completion checker.
In fractional units, the four reported rates satisfy
\begin{equation}
\label{eq:memory-metric-identity}
\begin{aligned}
\mathrm{Retrieval}&=\frac{n_R}{N_R},\quad
\mathrm{Gate}=\frac{n_G}{N},\\
\mathrm{Exec.}\mid\mathrm{Gate}&=\frac{n_{G\cap T}}{n_G}\quad(n_G>0),\\
\mathrm{Joint}&=\frac{n_{G\cap T}}{N}\\
&=\mathrm{Gate}\times(\mathrm{Exec.}\mid\mathrm{Gate}).
\end{aligned}
\end{equation}
Joint is the Memory score reported in Table~\ref{tab:main_results}.
Retrieval uses a narrower check and a different task subset, so it cannot
replace Gate in this identity. The identity holds per model before
rounding; it does not require the product of the cross-model means in
Figure~\ref{fig:capability-diagnosis-full}(b) to equal mean Joint.

\paragraph{Ordered skill routing in SOR.}
Let $\mathcal{P}_q$ be the set of acceptable skill routes for query $q$,
where each route $P\in\mathcal{P}_q$ is a set of skills that must all be
loaded to satisfy that route. For setting
$z\in\{\mathrm{SOR},\mathrm{MTSR}\}$, let $\ell_{q,z}(s)$ be the first
recorded load position of skill $s$, and let $b_{q,z}$ be the position of the
first non-skill tool call. Missing
loads and the absence of a non-skill call are assigned position $+\infty$.
The ordered-route check is
\begin{equation}
    g_{\mathrm{skill}}(q,z)
    =
    \mathbf{1}\!\left[
        \exists P\in\mathcal{P}_q:
        \max_{s\in P}\ell_{q,z}(s)<b_{q,z}
    \right].
\end{equation}
In SOR, this check is mandatory: at least one complete route must be loaded before any non-skill
tool call; the skills within a route need not be loaded in a prescribed
order, and every alternative route need not be loaded. Non-skill calls
include read-only and failed calls, not only successful state changes.
A load is recorded when the harness resolves a skill-loading call to an
available skill. Merely mentioning a skill or loading it after a business
call does not satisfy the gate. When available, call sequence indices
determine order; historical records containing only turn indices use the
evaluator's within-turn skill-first convention. Missing ordering evidence
does not pass.

\paragraph{Setting-specific Skill success.}
Let $v_{q,z}$ denote execution validity and $j_{1,q,z},j_{2,q,z}$ the two
judges' task-completion verdicts for the trajectory in setting $z$.
Define completion as
\begin{equation}
    T_z(q)=v_{q,z}\land j_{1,q,z}\land j_{2,q,z}.
\end{equation}
The success criterion is
\begin{equation}
    \operatorname{Succ}_z(q)=
    \begin{cases}
        g_{\mathrm{skill}}(q,z)\land T_z(q), & z=\mathrm{SOR},\\
        T_z(q), & z=\mathrm{MTSR}.
    \end{cases}
    \label{eq:skill-setting-success}
\end{equation}
Execution-validity checks reject model errors, unparseable business-tool
arguments, calls outside the effective tool set, and empty trajectories.
An ordinary tool failure is not by itself an automatic rejection:
the judges assess whether subsequent recovery, clarification, or refusal
appropriately addresses the task. MTSR rubrics assess the requested outcome
and task-declared constraints, not whether the planner loads an annotated
skill, follows its route order, or reproduces a reference trajectory.
Calling available tools directly, using loaded skills, or combining both
are all valid paths if the task is completed. Loading a skill alone is
never sufficient, and omission of skill loading alone is not an MTSR failure.

For diagnostics, Route reports $g_{\mathrm{skill}}$, Exec.$\mid$Route
reports completion conditional on passing that check, and Joint reports
their conjunction. Joint equals SOR success, but is only an ordered-route
diagnostic in MTSR, not its success rate. A failed ordered-route check
does not by itself establish that no skill was used.

\paragraph{Skill score aggregation.}
Let $\mathcal{Q}$ be the shared set of $N=200$ Skill queries. We compute
each setting's success rate on the full query set and retain their equal
average as the Skills score:
\begin{equation}
    \begin{aligned}
    S_z &= \frac{1}{N}\sum_{q\in\mathcal{Q}}\operatorname{Succ}_z(q),\\
    S_{\mathrm{Skills}} &= \frac{S_{\mathrm{SOR}}+S_{\mathrm{MTSR}}}{2}.
    \end{aligned}
    \label{eq:skill-score-aggregation}
\end{equation}
Scores are displayed as percentages. Missing or invalid predictions remain
in the denominator as failures. This aggregate measures completion under
two resource configurations; an MTSR success is not necessarily evidence
of skill use. The reported Skills scores use these setting-specific success
rates, as described in Appendix~\ref{app:evaluation-protocol}.